# Developing and Defeating Adversarial Examples

Computer Science and Engineering, Swarthmore College

Ian McDiarmid-Sterling
[imcdiar1@swarthmore.edu]

Professor Allan Moser
[amoser2@swarthmore.edu]

**Abstract**

*Breakthroughs in machine learning have resulted in state-of-the-art deep neural networks (DNNs) performing classification tasks in safety-critical applications. Recent research has demonstrated that DNNs can be attacked through adversarial examples, which are small perturbations to input data that cause the DNN to misclassify objects. The proliferation of DNNs raises important safety concerns about designing systems that are robust to adversarial examples. In this work we develop adversarial examples to attack the Yolo V3 object detector [1] and then study strategies to detect and neutralize these examples. Python code for this project is available at*
[https://github.com/ianmcdiarmidsterling/adversarial](https://github.com/ianmcdiarmidsterling/adversarial)

### 1.Introduction

This paper is organized as follows. In Section 2 we review prior work on adversarial examples and designing DNNs that are robust to adversarial examples. In Section 3 we present our procedure for developing adversarial examples and in Section 4 we examine strategies for detecting and neutralizing these examples. Discussion and conclusions follow in Sections 5 and 6.

There are two broad categories of adversarial attacks - targeted and untargeted. In a targeted adversarial attack, the goal of an attacker is to generate an attack that causes the system to produce a specific (targeted) incorrect classification. In an untargeted attack, the goal of an attacker is to generate any incorrect classification. Within these two categories adversarial attacks can be further categorized as 'general' or 'patch based'. In a general attack, changes are minute, but occur across an entire image, while in a patch based attack, a small region of the image is completely obscured by a patch. Most adversarial examples are not physically realizable, but research teams have recently started developing patch based adversarial examples in the real world.

In this work, we design patch based, untargeted adversarial examples because they are physically realizable in the wild.

Attempts have been made to make DNNs robust to adversarial examples by explicitly including the examples in the training data. We point out several weaknesses with these approaches in the next section, but the most serious, in our opinion, is that it may make it more difficult to detect the presence of an attack. The knowledge that a DNN is under attack is essential for designing appropriate mitigation/response strategies. Several strategies for detecting adversarial examples rely on 'white-box' access to the DNN. In this work we develop 'black box' strategies, where the DNN must be treated as an un-examinable process.

Other research appears promising. Rouani et al. [5] proposed a method for identifying adversarial examples by mapping out the feature space of normal data and adversarial examples. Their results demonstrate the power of this strategy, but their method is impossible in a black box scenario.

In this work we develop what to the best of our knowledge is a novel strategy for defending against adversarial examples. An ideal strategy has the following characteristics:
1. Effective in a black box scenario
2. Low computational time for adversarial example detection
3. Enables adversarial attack detection and neutralization
4. Robust to a wide range of adversarial examples

## 2. Background and Related Work

The work presented in [2,3,4] has focused on designing DNNs that are adversarially robust by explicitly training on adversarial examples. This approach has two main weaknesses. Firstly, the range of adversarial examples for which the DNN is robust directly depends on the range of training data. Secondly, as previously mentioned, the resulting DNN may be more robust to certain adversarial examples, but is unable to perform adversarial classification. This prevents attack recognition, inhibiting knowledge of when an attack has occurred.

Several researchers have demonstrated the effectiveness of adversarial examples against state-of-the-art object detection systems. For example, Eykholt et al. [6] designed a physically realizable adversarial patch that fools an object detection system into misclassifying an image of a stop sign as a speed limit sign (*Figure 1*).

Similarly, Thys et al. [7] demonstrated a physically realizable adversarial patch that is capable of completely suppressing the detection of a person in the image provided the patch is located near the person. Building on this work, Lee et al. [8] developed an adversarial patch that completely suppresses detections of multiple people, and can be located anywhere in the image.

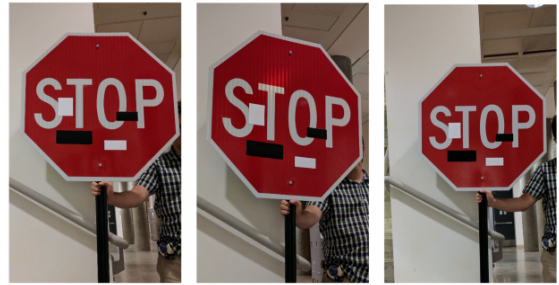

*Figure 1: The adversarial patches placed on a stop sign from [6] resulting in the complete misclassification of the stop sign as a speed limit sign.*

Inspired by the work of Lee et al. [8], we chose to develop adversarial examples for the Yolo V3 (henceforth Yolo) object detection model [1].

Yolo is a single shot object detection system capable of detecting 80 different object classes. In this work we chose to attack the 'person' class, but our approach works generally for any object class. In the sections that follow we denote the Yolo class label by $\hat{y}$ and indicate the dependence of the probability of observing this label on the corresponding input image, $x$, as $\mathbb{P}[\hat{y} \mid x]$.

Yolo divides an input image into an *n x m* grid of cells and for each cell the detector returns a list of bounding boxes (*B*), their associated confidence scores (*C*), and class probabilities (*CL*). Yolo then uses non-maximum suppression [9] to suppress extraneous detections and only return one (*B*,*C*,*CL*) triplet per object. When *C < 0.3* the associated object bounding box and class label are not returned and we say that the detector has failed to detect the object

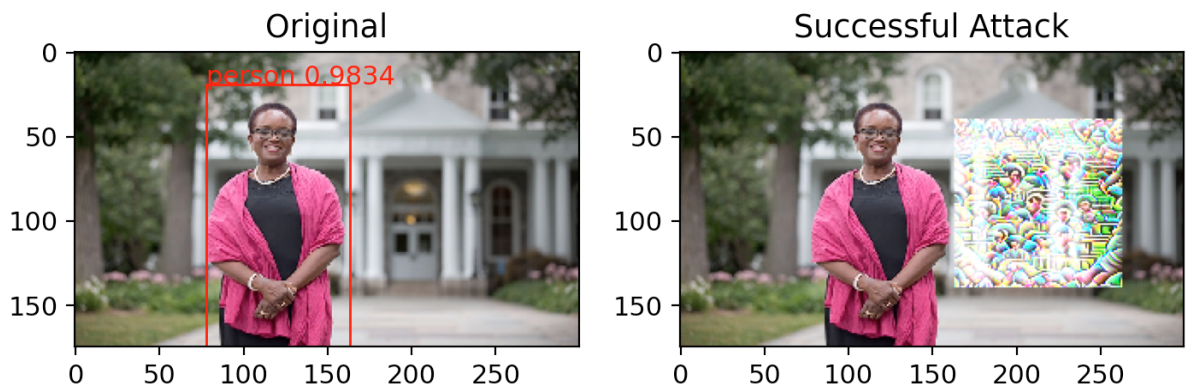

*Figure 2: Original image of the President of Swarthmore College, Dr. Valerie Smith (C=0.9834)(L). Image with the patch applied (C=0)(R).*

### 3. Patch development

As a measure of effectiveness, we computed the average Yolo confidence across all the images in the test dataset. We refer to this simply as 'average confidence' in the subsequent text.

In development of our adversarial examples, we created the most versatile patches possible by designing them to be both location and scene invariant.

To make the patches less noticeable, we minimized patch size given the average confidence was *< 0.3*. As shown in *Table 1* a *100x100* pixel patch was the smallest patch to meet this criterion. Note that in the absence of an adversarial example, Yolo detects the person with an extremely high average confidence over both the training and test datasets.

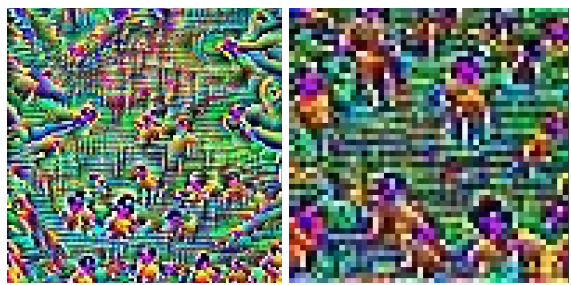

*Figure 3: Final versions of a trained patch: 100x100 pixels (R) and 50x50 pixels (L).*

| Patch Size (Pixels) | Mean training set confidence (32 images) | Mean test set confidence (3 images) |
|---|---|---|
| No Patch | 0.904 | 0.978 |
| 20x20 | 0.882 | 0.973 |
| 50x50 | 0.711 | 0.880 |
| 70x70 | 0.657 | 0.632 |
| 100x100 | 0.189 | 0.297 |
| 150x150 | 0.174 | 0.228 |

*Table 1: Mean confidence of patches trained for 1000 iterations with fixed hyper-parameters when applied to datasets.*

As discussed in [8,10,11] we implemented a variation of an expectation over transformation (EOT) operator. In our EOT operator, we randomized the location of the adversarial patch on the image to ensure patch performance was location invariant.

Similar to the notation in [10] let:

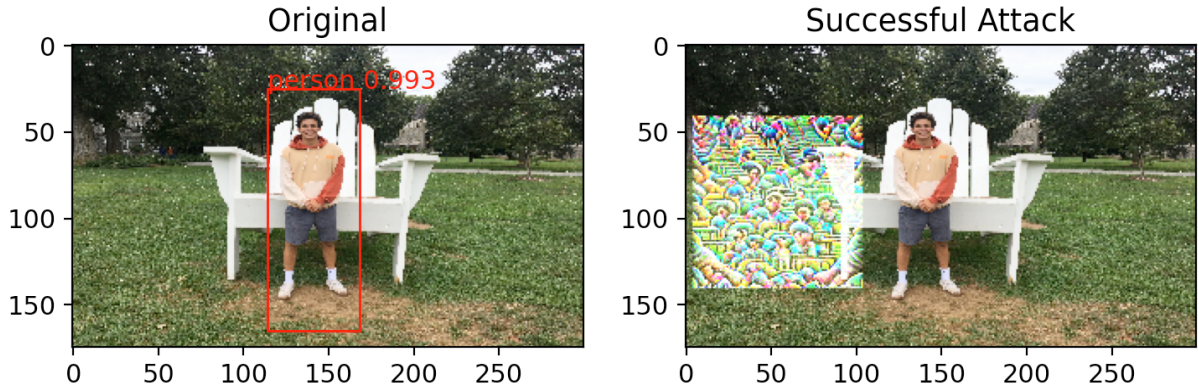

*Figure 4: Original image of the author in front of the giant Swarthmore adirondack lawn chair (C=0.993)(L). Image with the patch applied (C=0)(R).*

- $x \in (0,255)^{M \times N}$ denote the original 8-bit image
- $z \in R^{n \times n}$ denote a trainable patch
- $\tau$ from $R^{n \times n} \to R^{M \times N}$ denote a family of allowable affine transformations
- $m_\tau \in \{0,1\}^{M \times N}$ denote a binary mask where $\tau \in [1,M] \times [1,N]$
- $i,j$ denote integer row and column indices in the original image

We defined the patch application function:

$$a_t(x,z) = x + m_\tau \odot z$$

where $\odot$ represents the 'pixel wise' Hadamard product. We will denote this combination of the patch and original image as 'perturbed image'

We generate our adversarial patch by solving the following unconstrained minimization problem:

$$\hat{z} = \mathrm{argmin}_{z' \in Rn}(E_{t \sim T}[(\mathbb{P}[\hat{y}='person'|a_t(x,z')])^2])$$

We implemented the translation operation by creating a binary mask $m_{i,j}$ with ones for rows in the range $(i:i+n)$ and columns in the range $(j:j+n)$. Such a mask is easily constructed from the outer product of two binary vectors with blocks of ones in the appropriate components.

To solve the minimization problem above, we used the Fast Gradient Sign Method (FGSM) first introduced by Goodfellow et al. [3]. Given patch $z$ at time $t-1$ we compute $z$ at time $t$ using :

$$z_t = z_{t-1} + \varepsilon * sign(\nabla z_{t-1} J(\theta, z_{t-1}, y)) .$$

where $\varepsilon$ is the magnitude of the perturbation (learning rate). We clip pixel values in the patch so that $0 <= z_t <= 255$.

We observed that when a patch was placed along the edges of an image, the average confidence of the perturbed image was 0.801 higher than when the patch was placed in the interior regions of an image. We believe that this is a result of the image segmentation performed by Yolo. Specifically, if the patch does not intersect any of the image segments that contain an object (in this case a person), the patch will be unable to impact the detection of the object.

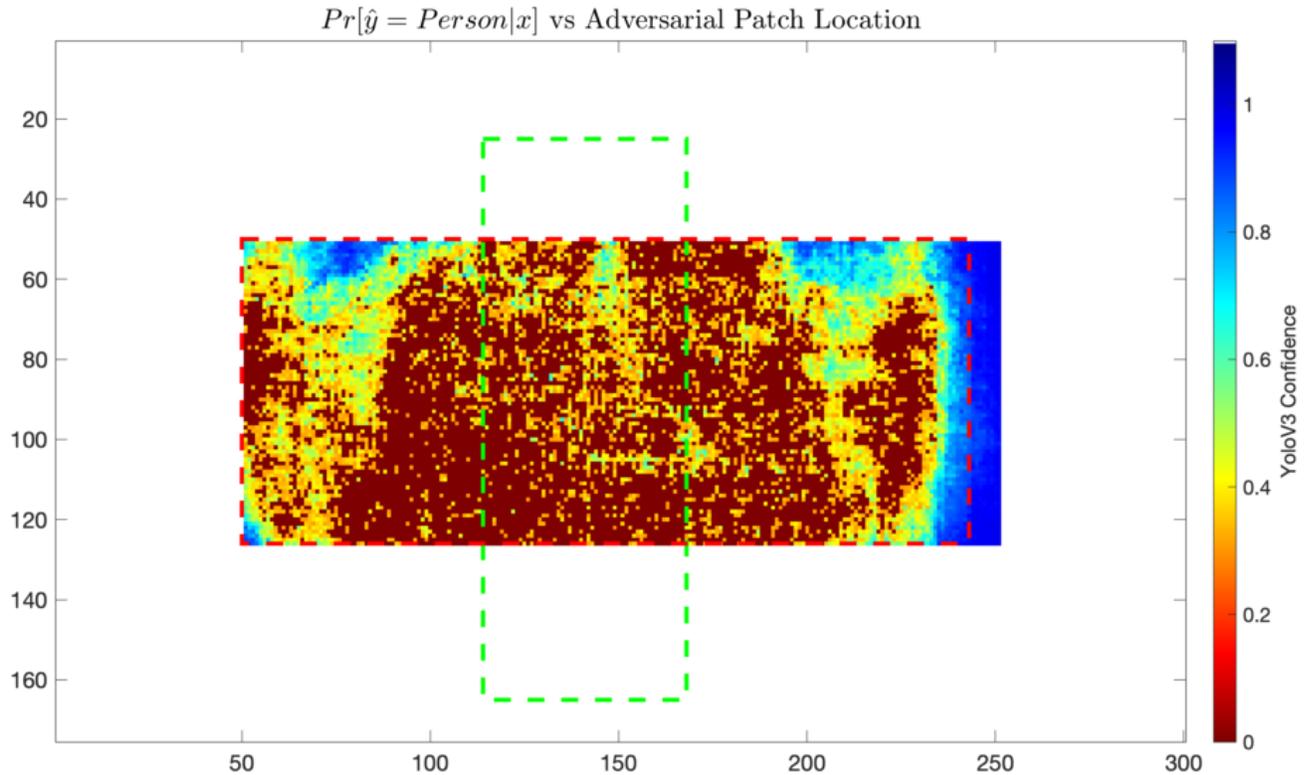

*Figure 5: Heatmap of mean Yolo confidence for the original image of the author (Figure 4), with each point representing the center of the 100x100 patch. The dashed green box represents the object (person) bounding box and the dashed red box represents possible patch locations with offset <=25. The white border region is the area in which patches cannot be placed without exceeding the image dimensions.*

To increase the chance that the patch would intersect one of the segments containing the person, we constrained the random patch locations by an offset, so that the edge of the patch was separated from the object by *<= 25* pixels (*Figure 5*). Within this constraint, at the vast majority of patch locations *C= 0* (*Figure 5*).

During patch training, we implemented a momentum calculation. The Gradient of *z* at time t can be expressed in terms of the gradient at the previous time step as:

$$\nabla z_t = a \nabla z_t + (1-a) \nabla z_{t-1}$$

Where alpha $\in$ *[0,1]*

Hyper-parameter optimization was used to select suitable values for initial learning rate and number of random transformations. To do this, we swept the range of acceptable hyper-parameters, training patches with every possible combination of parameters and comparing average confidence values. This revealed that the average confidence was independent of the initial learning rate for learning rates $\in$ *[0.5,20]*, when controlled for computational effort. Also, for fixed learning rates, confidence decreased as the number of random transformations increased, over the range from *5* to *1000*.

To ensure that the patch was more effective than a patch chosen by chance, we compared the trained patch to a patch of equivalent size with randomly generated pixel values. This 'noise patch' decreased average confidence by *< 0.010* (*Table 2*) compared to the actual trained patch which decreased average confidence by *0.692*.

| Type of Image | Mean training set confidence (32 images) | Mean test set confidence (3 images) |
|---|---|---|
| Original image | *0.904* | *0.978* |
| Image + 'Noise Patch' | *0.901* | *0.989* |
| Perturbed image | *0.189* | *0.297* |

*Table 2: Mean Yolo confidence for original images, images + 'noise patches', and perturbed images.*

### 4. Neutralizing the adversarial example

Once we generated an adversarial example, we evaluated its robustness with respect to two simple perturbations: additive Gaussian noise, with mean $\mu$ and standard deviation $\sigma$, and a Gaussian blur.

To properly assess the range of Gaussian noise, we studied individual perturbations with $\mu \in$ *[0,50]* and $\sigma \in$ *[1,100]*. For $\mu$ in the range, while $\sigma = c$, the average confidence of a perturbed image decreased by *< 0.052*. For $\sigma$ in the range, while $\mu = c$, the average confidence of a perturbed image decreased by *< 0.066*. These results suggest that adding Gaussian random noise is an ineffective method for defeating adversarial examples (*Table 3*).

| Type of image | Mean training set confidence (32 images) | Mean test set confidence (3 images) |
|---|---|---|
| Unmodified | *0.904* | *0.978* |
| Image + noise | *0.895* | *0.973* |
| Perturbed image | *0.189* | *0.297* |
| Perturbed image + noise | *0.187* | *0.153* |

*Table 3: Mean Yolo confidence for original images and perturbed images with and without additive Gaussian noise ($\mu$ =0, $\sigma$ =10)*

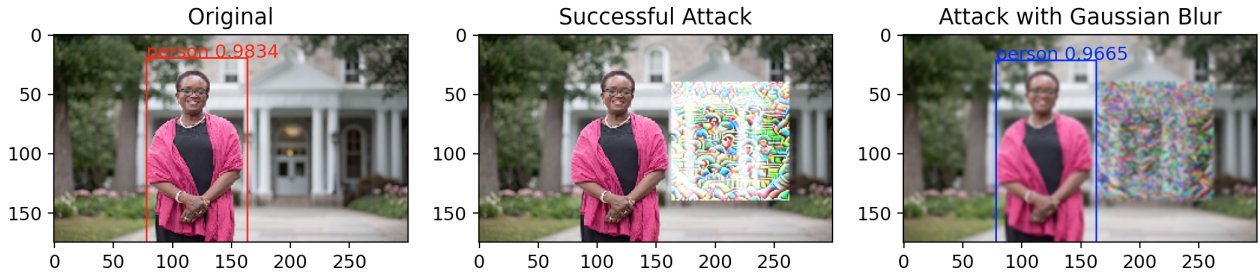

*Figure 6: Images of Dr. Smith: original(C=0.98)(L), perturbed(C=0)(M), perturbed with 3x3 Gaussian blur applied (C=0.97)(R).*

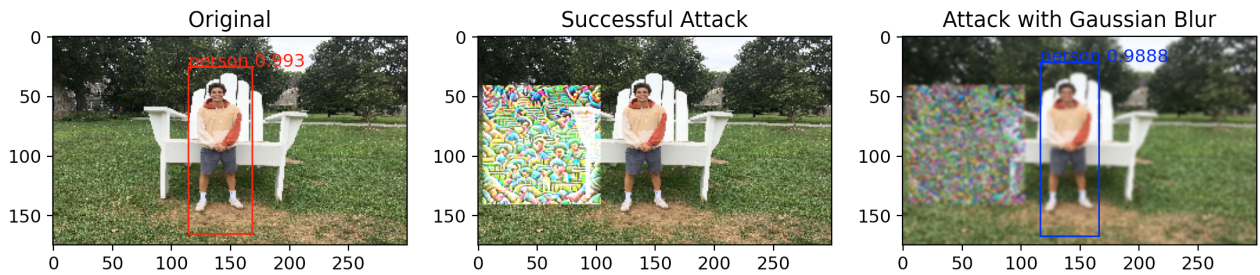

*Figure 7: Images of the author: original(C=0.99)(L), perturbed (C=0.0)(M), perturbed with 3x3 Gaussian blur applied (C=0.99)(R).*

Based on these observations, we hypothesized that the specific spatial structure in the patch is the key to suppressing detections. To test this theory, we applied a Gaussain blur [12] with a *3x3* convolution kernel in an attempt to disrupt this spatial structure *(Figures 6 and 7)*.

This was effective as *3x3* Gaussian blur increased the average confidence of a perturbed image by *0.628*, while only decreasing the average confidence of unmodified image by *< 0.010 (Table 4)*.

| *Type of image* | Mean training set confidence (32 images) | Mean test set confidence (3 images) |
|---|---|---|
| Unmodified | *0.904* | *0.978* |
| Image + blur | *0.904* | *0.969* |
| Perturbed image | *0.189* | *0.297* |
| Perturbed image + blur | *0.817* | *0.937* |

*Table 4: Mean Yolo confidence for original and perturbed images with and without Gaussian 3x3 blur.*

## 5. Discussion

The next phase of this research will consider the effects of incorporating blur and other 'trivial' countermeasures into the EOT operator, to explore the potential for generating patches that are robust to simple countermeasures like blur, lighting changes, and perspective shifts. In addition the next stage of this work will include generating more diverse patches that can suppress multiple detections in an image. Finally, we will explore adding a term to the loss function to incentivise similar values in adjacent pixels (a printability term), encouraging examples that are easier to realize physically.

One way to use simple filtering operations like Gaussian blur to detect adversarial examples is to use the difference in confidence between the original and blurred image as a predictor of the presence of an adversarial example.

## 6. Conclusion

In this work we developed an effective adversarial example for the Yolo object detector and then explored two simple strategies for defending against such examples.

First we created adversarial examples in the form of location invariant patches that when applied to images containing a person, decreased the average confidence of the Yolo object detector by *0.681*, relative to the confidence on the unmodified images.

Then we explored two simple strategies to defeat the adversarial examples: applying additive Gaussian noise or a Gaussian blur operator. These strategies do not require access to or knowledge of the Yolo architecture, and can be performed relatively quickly (compared to computing a forward pass with Yolo). Of the two operators, the Gaussian blur operator allowed detection and neutralization of the adversarial examples, whereas we found additive Gaussian noise to be ineffective. When we augmented a perturbed image with *3x3* Gaussian blur, average confidence *increased* by *0.640*, relative to the confidence in a perturbed image without Gaussian blur. We show examples of specific images where *3x3* Gaussain blur restores an accurate prediction with confidence *> 0.9*. In future work we plan to study the versatility of additional filtering strategies for detecting and neutralizing adversarial examples.

### 6.1. Acknowledgments

We thank our reviewers for their constructive feedback. Specifically we thank Dr. David Sterling for his guidance, suggestions, feedback, and support. This work was funded by a Swarthmore College Summer Research Fellowship grant (SOAR-NSE).